\documentclass[letterpaper, 10pt, journal, twoside]{IEEEtran}  

\usepackage{amsmath} 
\usepackage{amssymb}  

\usepackage{orcidlink}
\hypersetup{
    colorlinks=true,
    linkcolor=blue,
    citecolor=blue,
    urlcolor=black
}

\usepackage{tikz}
\usepackage{pgfplots}
\pgfplotsset{compat=1.18}
\usetikzlibrary{external}
\usetikzlibrary{matrix, decorations.pathreplacing, calc, positioning, fit, fadings, shapes.arrows, arrows.meta, shapes.geometric, backgrounds}
\usetikzlibrary{shadings}
\usepackage{tikzscale}

\usepackage{algorithm}
\usepackage{algorithmicx}
\usepackage[noend]{algpseudocode}
\usepackage[T1]{fontenc}
\usepackage{svg}
\usepackage{xspace}
\usepackage{siunitx}

\newcommand{\ours}{SOFOF\xspace}

\newcommand\copyrighttext{\footnotesize \textcopyright~2025 IEEE. Personal use of this material is permitted. Permission from IEEE must be obtained for all other uses, in any current or future media, including reprinting/republishing this material for advertising or promotional purposes, creating new collective works, for resale or redistribution to servers or lists, or reuse of any copyrighted component of this work in other works.%
DOI: \href{https://ieeexplore.ieee.org/document/10945369}{10.1109/LRA.2025.3555939}
}

\newcommand\copyrightnotice{%
    \begin{tikzpicture}[remember picture,overlay]%
     \node[%
        anchor=south, %
        yshift=5pt%
    ] at (current page.south)%
     {\fbox{\parbox{\dimexpr\textwidth-\fboxsep-\fboxrule\relax}{\copyrighttext}}};%
     \end{tikzpicture}%
}

\begin{document}

\title{A Generic Service-Oriented Function Offloading Framework for Connected Automated Vehicles}

\author{Robin Dehler$^{1}$\orcidlink{0000-0002-1999-9286} and Michael Buchholz$^{1}$\orcidlink{0000-0001-5973-0794}

\thanks{Manuscript received: 10 January 2025; Accepted 11 March 2025.}
\thanks{This paper was recommended for publication by Editor M. Ani Hsieh upon evaluation of the Associate Editor and Reviewers' comments.}%
\thanks{This work has been financially supported by the Federal Ministry of Education and Research (project autotech.agil, FKZ 01IS22088W). }
\thanks{$^{1}$The authors are with the Institute of Measurement, Control and Microtechnology, Ulm University, D-89081 Ulm, Germany.\\E-mail addresses: \{firstname\}.\{lastname\}@uni-ulm.de}
}

\markboth{IEEE Robotics and Automation Letters. Preprint Version. Accepted March, 2025}
{Dehler \MakeLowercase{\textit{et al.}}: A Generic Service-Oriented Function Offloading Framework for Connected Automated Vehicles} 

\maketitle

\begin{abstract}
Function offloading is a promising solution to address limitations concerning computational capacity and available energy of Connected Automated Vehicles~(CAVs) or other autonomous robots by distributing computational tasks between local and remote computing devices in form of distributed services.
This paper presents a generic function offloading framework that can be used to offload an arbitrary set of computational tasks with a focus on autonomous driving.
To provide flexibility, the function offloading framework is designed to incorporate different offloading decision making algorithms and quality of service~(QoS) requirements that can be adjusted to different scenarios or the objectives of the CAVs.
With a focus on the applicability, we propose an efficient location-based approach, where the decision whether tasks are processed locally or remotely depends on the location of the CAV.
We apply the proposed framework on the use case of service-oriented trajectory planning, where we offload the trajectory planning task of CAVs to a Multi-Access Edge Computing~(MEC) server.
The evaluation is conducted in both simulation and real-world application.
It demonstrates the potential of the function offloading framework to guarantee the QoS for trajectory planning while improving the computational efficiency of the CAVs.
Moreover, the simulation results also show the adaptability of the framework to diverse scenarios involving simultaneous offloading requests from multiple CAVs.
\end{abstract}

\copyrightnotice
\begin{IEEEkeywords}
Intelligent Transportation Systems, Distributed Robot Systems, Networked Robots
\end{IEEEkeywords}

\section{Introduction}
\IEEEPARstart{F}{or} a reliable and safe application, many functions of autonomous systems require elaborate algorithms with high demands on hardware usage and energy consumption.
Due to the often dynamic nature of autonomous systems, e.g., autonomous electric vehicles, both the available hardware and battery are limited.
However, the functions of connected automated vehicles (CAVs) are nowadays no longer limited exclusively to hardware and software within the vehicles, but also include both hardware and software components that are employed on external computational units, like servers.

Function offloading is one of the common approaches to use the extended capabilities of such hardware by a CAV.
Empowered through the development of wireless communication technologies, e.g., vehicle-to-anything~(V2X), CAVs can offload specific functions with the goal to decrease their own overall workload or energy consumption.
A wide variety of functions can be considered for function offloading, ranging from infotainment and comfort applications with lower quality of service (QoS) demands, e.g., automatic air conditioning control systems~\cite{10.1007}, to safety-critical autonomous driving functions with higher QoS demands.

For function offloading, different computing technologies were developed, two prominent ones being Mobile Cloud Computing~(MCC) and Multi-Access Edge Computing~(MEC).
MCC is the technology of using powerful cloud servers as additional hardware.
However, MCC has the disadvantage of highly variable latency in the communication with the autonomous system, which limits its application for the often latency-sensitive functions of CAVs~\cite{wang_20}.
MEC provides an alternative to MCC to overcome its limitations by employing servers at the edge of the network with better performance regarding latency.
Consequently, MEC is an important technology for function offloading.

Service-orientation is another trend for autonomous driving and complex systems in general.
For service-oriented architectures~(SOAs), each function of a CAV can be considered as a service that can easily be modified or exchanged without the need to change the entire computational stack.
The concept of SOA is very promising, as it offers the software flexibility needed for autonomous driving~\cite{8916841}.
In combination, the two technologies function offloading and SOA provide a generic framework for autonomous driving.

\begin{figure}[t]
    \centering
    \includegraphics[width=\linewidth]{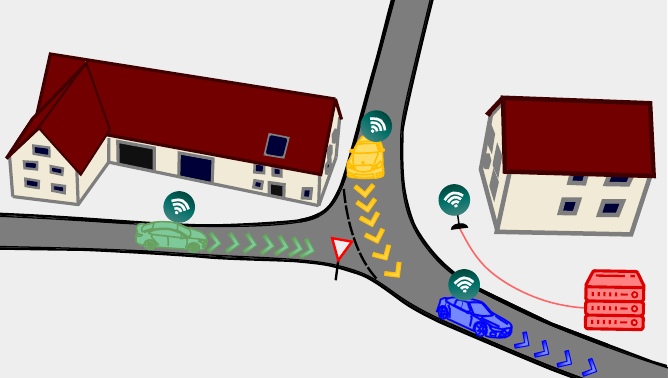}
    \caption{Considered intersection scenario for service-oriented function offloading including multiple CAVs and a MEC server.}
    \label{scenario}
\end{figure}

A lot of research on function offloading focuses on algorithms for function offloading decision making and resource allocation, with the objective to decide which functions should be offloaded from autonomous devices to the servers.
The function offloading decision making and resource allocation problem is commonly described as an optimization problem considering, e.g., transmit power, bandwidth, computational capacities, energy consumption, and delay requirements~\cite{8240666}.
In contrast to the literature, this work aims to provide an overall framework for function offloading that includes the required management functionalities to offload tasks or functions.
The function offloading decision making algorithms from the literature can, e.g., be integrated as one part into the proposed framework.
Further aspects of function offloading, such as schedulability, security, and fault-tolerance analysis, are outside the scope of this paper and will be considered in future work.

We consider the distribution of a function on different computers as different services that, in the concept of SOA, are easily adaptable and interchangeable by using the framework.
The proposed Service-Oriented Function Offloading Framework~(\ours) comprises both instances of function offloading, i.e., service providers that provide services for others and service requesters that request to offload tasks.
In combination, the instances for service providers and requesters are responsible for managing the state transitions of services on multiple devices.
Thus, if not used within an existing SOA, the framework can also operate as the service orchestrator within the concept of the automotive service-oriented architecture (ASOA)~\cite{8916841}, otherwise, the respective functionalities can also be integrated into the existing orchestrator.
In the following, we will concentrate on \ours and its special functionalities and refer to, e.g., \cite{8916841}, for more detailed discussions on the ASOA.
Using the framework can improve the computation of CAVs by increasing the number of services that can be used for autonomous driving tasks.

For the evaluation, we consider a scenario at a real-world suburban traffic junction with a deployed MEC server, as seen in Fig.~\ref{scenario}.
We apply and evaluate the framework for the use case of service-oriented trajectory planning in both simulation and real-world application on our test site~\cite{buchholz_22}.

Our main contributions are as follows:
\begin{itemize}
    \item With \ours, we present a generic framework that can be used by both function offloading instances, i.e., service provider and service requester.
    \item We propose an efficient location-based function offloading decision making algorithm, where the decision depends on the location of the CAV and an offloading area dedicated to the server.
    \item We analyze \ours on the application of the use case of service-oriented trajectory planning.
    \item We extensively evaluate the framework with the use case of service-oriented trajectory planning in both simulation and our real-world autonomous driving test vehicle, showing the applicability of the approach especially for real-world scenarios.
\end{itemize}
\section{Related Work}
The research of function offloading mainly addresses the optimization of computational efficiency and resource management in distributed systems.
To solve the resulting optimization problems, most approaches propose solutions involving decision making and resource allocation algorithms.
The most prominent approaches can be categorized into Mixed-Integer Programming~(MIP) approaches, Lyapunov optimization techniques, and approaches using Machine Learning~(ML)~\cite{SAEIK2021108177, acheampong-22}.

When addressed as a MIP, the optimization problem is generally NP-hard, scaling with the number of remote agents, e.g., autonomous vehicles~\cite{7553459}.
Thus, the problem is often solved using game theory~\cite{8745530, 9200665}.
Applying game theory, vehicles are players competing for the available resources while considering additional aspects, e.g., a maximum latency.
The main advantages of Lyapunov optimization techniques are direct stability analysis and relatively low algorithm complexity~\cite{SAEIK2021108177}.
However, choosing an appropriate Lyapunov function is not trivial.
Many ML concepts describe the offloading problem as a Markov Decision Process~(MDP) and use Reinforcement Learning~(RL) for making offloading decisions~\cite{9725258, 9161406, 9989360, 9831431, 8771176}.
Since each concept provides different advantages, they are often combined to benefit from them while reducing disadvantages, e.g., ~\cite{9174795, 9449944, 9166745}.
Note that all of the presented approaches evaluate the proposed offloading algorithm in simulation without real-world analysis.

Another research focus in the area of distributed systems for autonomous driving addresses the usage of different services with a focus on SOAs.
It has been established that the concept of SOA is suitable for meeting the requirements of CAVs~\cite{7930217}.
A version of a SOA specifically for autonomous driving, called ASOA, was introduced in~\cite{8916841}.

Since distributed systems pose greater challenges on safety and security, safety and security analysis is of research interest~\cite{9285284, 7930217, 5281251}. 
In~\cite{9740161}, the service of vehicle collaboration is used in combination with a digital twin for collaborative and distributed automated driving.
Another approach for cooperative driving in critical sections is proposed in~\cite{7725527}.
In \cite{10.1007}, the concept of SOA is extended to cloud architectures and specific requirements are analyzed.
The authors propose an orchestrator design that manages the execution of both local and cloud services.

Similarly to \cite{10.1007}, this work combines function offloading and the concept of SOA in a way that the proposed framework can be employed to use different distributed services for connected automated driving.
Compared to the state of the art, we approach the function offloading task in a more general perspective, combining different components, such as function offloading decision making, resource allocation, service orchestration, and QoS considerations into one framework.
The service-oriented approach allows flexible interchangeability of different components such that existing algorithms, e.g., for function offloading decision making, can be easily integrated into the framework.

\section{Function Offloading Framework}
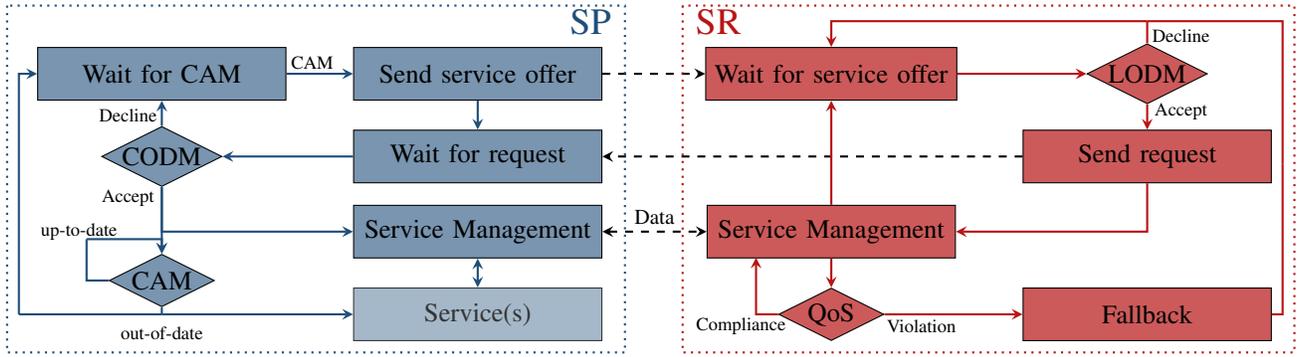
\begin{figure*}[t]
    \centering
    \tikzstyle{arrow} = [thick,->,>=stealth]
\tikzstyle{box} = [rectangle, minimum height=.7cm, minimum width=3.3cm]

\definecolor{darkblue}{HTML}{1f4e79}
\definecolor{salmon}{HTML}{ff9c6b}
\definecolor{maroon}{HTML}{b81414}

\begin{tikzpicture}
\def\lrshift{3.7cm}
\def\spacing{.1cm}
%
\node[draw, fill=darkblue!60, text=black, box, xshift=-2cm, yshift=-.2cm] (wait-cam) {Wait for CAM};
\node[right of=wait-cam, yshift=0.15cm, xshift=1cm, align=left] (cam-rx) {\scriptsize CAM};
\node[draw, fill=darkblue!60, box, right of=wait-cam, xshift=3.2cm] (send-rp) {Send service offer};

\node[draw, fill=maroon!70, box, right of=send-rp, xshift=\lrshift] (wait-offer) {Wait for service offer};
\node[draw, fill=maroon!70, diamond, aspect=2, right of=wait-offer, xshift=3.2cm, align=left, minimum height=.8cm, minimum width=1.6cm] (dodm) {};
\node[right of=wait-offer, xshift=3.2cm, align=left] (dodm-2) {LODM};
\node[below of=dodm, xshift=.45cm, yshift=.5cm] {\scriptsize Accept};
\node[above of=dodm, xshift=0.45cm, yshift=-.5cm] {\scriptsize Decline};
\node[draw, fill=maroon!70, box, below of=dodm, yshift=-\spacing] (send-rr) {Send request};

\node[draw, fill=darkblue!60, box, below of=send-rp, yshift=-\spacing] (wait-r) {Wait for request};
\node[draw, fill=darkblue!60, diamond, aspect=2, left of=wait-r, xshift=-3.2cm, minimum height=.8cm, minimum width=1.6cm] (codm) {};
\node[left of=wait-r, xshift=-3.2cm] (codm-2) {CODM};
\node[below of=codm, xshift=-.45cm, yshift=.45cm] {\scriptsize Accept};
\node[above of=codm, xshift=-0.45cm, yshift=-.45cm] {\scriptsize Decline};
\node[draw, box, fill=darkblue!60, below of=codm, xshift=4.2cm] (service-man) {Service Management};
\node[draw, black!80, box, fill=darkblue!40, below of=service-man, yshift=-\spacing] (mec-service) {Service(s)};
\node[draw, diamond, aspect=2, fill=darkblue!60, below of=codm, yshift=-0.55cm-\spacing, minimum height=.7cm, minimum width=1.4cm] (cam-check) {};
\node (cam) at (cam-check){CAM};
\node[left of=cam-check, xshift=-.1cm, yshift=.65cm] {\scriptsize up-to-date};
\node[below of=cam-check, yshift=.3cm] {\scriptsize out-of-date};

\node[draw, fill=maroon!70, box, right of=service-man, xshift=\lrshift] (service-man-r) {Service Management};
\node[left of=service-man-r, xshift=-1.35cm, yshift=0.2cm] (data) {\footnotesize Data};

\node[draw, diamond, aspect=2, fill=maroon!70, below of=service-man-r, yshift=-\spacing, minimum height=.7cm, minimum width=1.4cm] (qos-check) {};
\node[below of=service-man-r, yshift=-\spacing] (qos) {QoS};
\node[draw, box, fill=maroon!70, right of=qos-check, xshift=3.2cm] (fallback) {Fallback};
\node[right of=qos-check, xshift=.2cm, yshift=-.15cm] {\scriptsize Violation};
\node[left of=qos-check, xshift=-.2cm, yshift=-.15cm] {\scriptsize Compliance};

\draw [arrow, darkblue] (wait-cam) -- (send-rp);
\draw [arrow, darkblue] (send-rp) -- (wait-r);
\node [below of=send-rp, yshift=.45cm] (help-r) {};
\draw [arrow, darkblue] (wait-r) -- (codm);
\draw [arrow, darkblue] (codm) -- (wait-cam);
\draw [arrow, darkblue] (codm) |- (service-man);
\draw [arrow, darkblue] (codm) -- (cam-check);
\node (cam-check-helper1) at (cam-check |- mec-service) {};
\node[left of=cam-check-helper1, xshift=-0.9cm] (cam-check-helper2) {};
\draw [thick, darkblue] (cam-check) -- (cam-check-helper1.center);
\draw [arrow, darkblue] (cam-check-helper1.center) -- (mec-service);
\draw [thick, darkblue] (cam-check-helper1.center) -- (cam-check-helper2.center);
\draw [arrow, darkblue] (cam-check-helper2.center) |- (wait-cam);
\node [left of=cam-check, yshift=.55cm] (cam-check-helper3) {};
\draw [thick, darkblue] (cam-check) -| (cam-check-helper3.center);
\draw [arrow, darkblue] (cam-check-helper3.center) -| (cam-check);

\draw [thick, <->, >=stealth, darkblue] (service-man) -- (mec-service) {};

\draw [arrow, dashed] (send-rp) -- (wait-offer);
\draw [arrow, dashed] (send-rr) -- (wait-r);
\draw [thick, <->,>=stealth, dashed] (service-man) -- (service-man-r);

\draw [arrow, maroon] (wait-offer) -- (dodm);
\draw [arrow, maroon] (dodm) -- (send-rr);
\node [above of=dodm, yshift=-.3cm] (decline-help) {};
\draw [thick, maroon] (dodm) -- (decline-help.center);
\draw [arrow, maroon] (decline-help.center) -| (wait-offer);
\draw [arrow, maroon] (send-rr) |- (service-man-r);
\draw [arrow, maroon] (service-man-r) -- (wait-offer);
\draw [arrow, maroon] (service-man-r) -- (qos-check);

\node[below of=service-man-r, xshift=-1cm, yshift=0.78cm] (qos-yes) {};
\draw [arrow, maroon] (qos-check) -| (qos-yes);
\draw [arrow, maroon] (qos-check) -- (fallback);
\node[right of=fallback, yshift=2cm, xshift=0.8cm] (fallback-help) {};
\draw[thick, maroon] (fallback) -| (fallback-help.center);
\draw[thick, maroon] (fallback-help.center) |- (decline-help.center);

\node [above of=send-rp, xshift=1.5cm, yshift=-.3cm, text=darkblue] {\Large SP};
\node [above of=wait-offer, xshift=-1.5cm, yshift=-.3cm, text=maroon] {\Large SR};
\draw[darkblue,thick,dotted] ($(wait-cam.north west)+(-0.4,0.55)$)  rectangle ($(mec-service.south east)+(0.3,-0.15)$);
\draw[maroon,thick,dotted] ($(wait-offer.north west)+(-0.3,0.55)$)  rectangle ($(fallback.south east)+(0.3,-0.15)$);

\end{tikzpicture}
    \caption{Overview of \ours, simplified for one SP and one SR. After an offloading request is denoted as successful with either the local (LODM) or centralized offloading decision making (CODM), the two service management instances handle the data flow for the offloaded services, respectively. A QoS check on the SR initiates a fallback if the QoS requirements are violated. The CAM check on the SP continuously checks the position of the CAV.}
    \label{framework}
\end{figure*}
The objective of function offloading is commonly described using an optimization function that consists of different components, e.g., the minimization of latency, the reduction of energy consumption, or the maximization of QoS.
Our approach to service-oriented function offloading focuses especially on the QoS and on the improvements that additional services on distributed hardware can provide compared to only local execution.
For that, we consider traffic scenarios in a suburban area with multiple CAVs and a single MEC server unit similar to Fig.~\ref{scenario}.
The CAVs are considered to continuously send Cooperative Awareness Messages~(CAMs) as they are defined by the European Telecommunications Standards Institute~(ETSI)~\cite{etsi_en_302_637-2_intelligent_2014}.

\ours comprises two distinct instances: It either serves as a service provider~(SP) or a service requester~(SR).
In its role as a SP, \ours handles incoming service requests and activates and manages services for accepted requests.
As a SR, its purpose is to analyze which tasks should be offloaded, request the respective services, and manage incoming data.
Figure~\ref{framework} shows the elements of \ours for both roles.
In the following, the two different operating modes are explained.

\subsection{Service Provider}
As shown on the left of Fig.~\ref{framework}, when operating as a SP, the first task is to wait for CAVs to communicate their position to the server, e.g., through CAMs. 
For clarity of the figure, the separate module of the CAV for the continuous sending of CAM messages is not visualized. 
When a CAV is detected by a received CAM, the SP signalizes its availability to the SR instance of the CAV in the form of a service offer message and waits for a request of the detected CAV.
After receiving a request through a particular request message from the SR, a centralized offloading decision making~(CODM) algorithm decides whether the request for the specific service(s) is accepted or rejected.
The generalized structure of our framework allows for a flexible interchangeability of the CODM algorithm.
If accepted, \ours executes each accepted service in a separate thread and manages the services through the service management.
The block for the activated service(s) for this SR are visualized in lighter color in Fig.~\ref{framework}, since it is not directly a component of the framework.

For some services, additional data from the served SR is required.
The service management accounts for that by transforming and forwarding the received data to the corresponding service in the required format.
The data generated by the services is received by the service management unit and forwarded through V2X communication to the SRs.

The described process chain is executed independently for each SR connected to the SP. However, dependent on the implementation, the CODM knows about all other ongoing requests and running offloading services, thus can take the current overall capacity utilization of the SR into account. 

\subsection{Service Requester}
When acting as a SR, \ours first waits for incoming service offers from SP instances.
Note that the continuous sending of the CAMs, which are necessary on the SP side, is done in parallel in a separate module, which is not shown to keep the figure clearly arranged.
When an incoming service offer is received, a local offloading decision making~(LODM) algorithm on the SR side evaluates the benefit of the provided services.
Similarly to the CODM, the LODM is easily exchangeable.
If the SR wants to use some of the provided services, a request message specifying the requested services is sent to the corresponding SP.

After sending a request, the service management waits for the incoming data for the respective services.
After the arrival of the first data for an offloaded function, the function is deactivated and the received data is used for further functions, e.g., vehicle control.

Some services have specific QoS requirements, e.g., maximum latency or minimum frequency of received data.
A QoS control instance can be employed to each offloaded service.
Thus, different QoS profiles can be adjusted to each service depending on the QoS requirements.
If the QoS requirements of a service are violated, a fallback is activated through the service management that stops the offloading of that particular service and simultaneously activates the service on the SR to ensure the QoS for the function, respectively.

\section{Service-Oriented Trajectory Planning}
\ours serves as a generic framework to handle and manage an arbitrary number of different services for function offloading.
In this paper, we focus on one use case in particular, i.e., service-oriented trajectory planning for CAVs.
The objective is to use function offloading to provide the service \textit{trajectory planning on the MEC server}~(MEC-TPL).
The SRs can request and use the service MEC-TPL as an addition to the service \textit{local trajectory planning}~(CAV-TPL).
Within this paper, we integrate our framework into our existing, traditional architecture for autonomous driving.
We only consider the function trajectory planning as a service, all other parts are classical functions onboard the vehicle. 
Within the concept of ASOA, our framework hence operates as the orchestrator for multiple trajectory planning services.

For our use case, we developed two algorithms for CODM and LODM that are explained in the following.

\subsection{Centralized Offloading Decision Making}
A CODM employed on the SP is used to accept or reject incoming service requests.
In this work, we use a location-based CODM algorithm that depends on two conditions.
The SP has stored maps and an offloading area in which function offloading is possible in the form of polygon points.
The offloading area can, e.g., be determined dependent on the network connection.
Within the offloading request, the SRs send the planned route to the SP.
Initially, the received route is compared with the maps available to the SP.
If a corresponding map is present, a second condition checks whether the route is long enough in the area where offloading is possible using the ray-casting algorithm~\cite{ROTH1982109}.
For the distance, the received route waypoints are analyzed until the first waypoint is outside the polygon.
Since the route information does not include vehicle speeds, a constant velocity (CV) model is used on the current speed of the SR.
The CV model serves as a simple analysis instead of a forward planning approach, e.g., considering road topology and velocity limits, based on the trade-off between computational complexity and performance of the CODM algorithm.
The time is taken from the sum of the Euclidean distances between successive waypoints until the last waypoint in the polygon area given the constant velocity $v_c$, see Eq.~(\ref{eq:codm}).

For the condition, a threshold value $t_{min}$ is used to decide whether offloading is worthwhile. Thus, the condition can be stated as
\begin{align}\label{eq:codm}
    t_{min} \leq \sum_{i=1}^l\frac{ \sqrt{(x_i-x_{i-1})^2+(y_i-y_{i-1})^2} }{ v_c },
\end{align}
with $x_i$ and $y_i$ being the $x$- and $y$-coordinates of waypoint $i$ and $l$ the index of the last waypoint in the polygon area.

While the service management handles the activated services, the latest CAM of the served CAV is constantly updated.
If the CAV leaves the offloading area or if the latest CAM is older than two seconds, the offloading is terminated and the activated services are deactivated.

\subsection{Local Offloading Decision Making}
A received service offer contains information about the position of the SP pos$_{\text{SP}}$.
A LODM algorithm is used to accept or decline the offloading offer.
We again use a location-based decision making process that is summarized in Alg.~\ref{alg:LODM}.
For the LODM, an arbitrary offloading radius $r_{\text{off}}$ can be set.
This radius determines the maximum distance that a SR must have to the respective connection point to the SP so that function offloading is possible.

Alg.~\ref{alg:LODM} iterates through the planned path of the CAV path$_\text{CAV}$ and compares the distance of the points to the SP connection point pos$_\text{SP}$ with $r_{\text{off}}$ (line 2).
As long as the path is within $r_{\text{off}}$, the distance d$_\text{pos}$ from the current position pos$_\text{curr}$ to the previous pos$_\text{pos}$ is added to $d_\text{passed}$ (lines 3 - 4).
As soon as $d_\text{passed}$ is greater than a minimum distance $d_\text{min}$, which means that path$_\text{CAV}$ is long enough within $r_\text{off}$ to pos$_\text{SP}$, the function offloading request is accepted (lines 5 - 6).

\begin{algorithm}[t]
    \caption{Local Offloading Decision Making}
    \label{alg:LODM}
    \hspace*{\algorithmicindent} \textbf{Input:} path$_\text{CAV}$, SR position pos$_\text{last}$, $r_{\text{off}}$, pos$_{\text{SP}}$, $d_{\text{min}}$ \\
    \hspace*{\algorithmicindent} \textbf{Output:} Bool value for decision 
    \begin{algorithmic}[1]
    \State $d_{\text{passed}} = 0$, pos$_{\text{curr}} =$ path$_\text{CAV}[1]$
    \While{Distance of pos$_{\text{curr}}$ to pos$_{\text{SP}}$ < $r_{\text{off}}$}
        \State $d_{\text{pos}} = $ Distance between pos$_{\text{curr}}$ and pos$_{\text{last}}$
        \State $d_{\text{passed}} \leftarrow d_{\text{passed}} + d_{\text{pos}}$
        \If{$d_{\text{passed}} > d_{\text{min}}$}
            \State \textbf{return} True
        \EndIf
        \State pos$_{\text{last}} \leftarrow$ pos$_{\text{curr}}$, pos$_{\text{curr}} \leftarrow$ next point of path$_\text{CAV}$
    \EndWhile
    \State \textbf{return} False
    \end{algorithmic}
\end{algorithm}

\subsection{Trajectory Planning}
We use common ETSI Intelligent Transportation Systems~(ITS) messages~\cite{etsi_es_202_663_intelligent_2009} for sending the data from the CAV to the MEC server and vice versa via a cellular network.
The particular trajectory planning algorithm for both the CAV-TPL and the MEC-TPL was introduced in~\cite{10186535}.
It uses, in a nutshell, an Augmented Lagrangian-Iterative Linear Quadratic Regulator (AL-ILQR) optimization method in a Model Predictive Control (MPC) fashion to compute trajectories.
Note that the trajectory planning algorithms for different services can also be different, e.g., using a more complex planner for the MEC-TPL to make use of the powerful hardware on the MEC server for calculating higher quality trajectories as an additional benefit for function offloading.

\begin{figure}[t]
    \centering
    \tikzstyle{arrow} = [thick,->,>=stealth]

\definecolor{darkblue}{HTML}{1f4e79}
\definecolor{salmon}{HTML}{ff9c6b}
\definecolor{maroon}{HTML}{b81414}

\tikzstyle{box} = [draw, rectangle, minimum width=1.9cm, minimum height=0.9cm, font=\footnotesize]

\begin{tikzpicture}
\node (cav) [maroon] {\scriptsize \textbf{CAV}};
\node (mec) [below of=cav, yshift=.5cm, darkblue]{\scriptsize \textbf{MEC}};
\node (v2x) [box, right of=cav, xshift=.3cm, yshift=.4cm, fill=maroon!80, text depth=0.15cm, align=center] {ETSI V2X};
\node (cpm) [below of=v2x, yshift=0.7cm, xshift=-0.45cm, fill=white, inner sep=1.5pt] {\scriptsize CPM};
\node (cam) [below of=v2x, yshift=0.7cm, xshift=0.45cm, fill=white, inner sep=1.5pt] {\scriptsize CAM};

\node [box, below of=v2x, yshift=-0.3cm, fill=darkblue!60, align=center] (env) {Environment\\ Model };

\node [box, right of=env, xshift=1.5cm, fill=darkblue!60, align=center] (plan) {MEC-TPL};

\node [box, right of=plan, xshift=1.5cm, fill=darkblue!60, align=center, text height=0.9cm, inner sep=0.1pt] (sc1) {Service\\ Management};
\node (mcm) [above of=sc1, yshift=-0.66cm, fill=white, inner sep=1.5pt] {\scriptsize MCM};

\node [box, fill=maroon!80, align=center] (sc2) at (sc1 |- v2x) {Service\\ Management};

\node[below of=cam, yshift=.35cm] (arrow-cam) {};
\draw [arrow] (cam) -| (plan);

\node[below of=cpm, yshift=.35cm] (arrow-cpm) {};
\draw [arrow] (cpm) -- (cpm |- env.north);

\draw[arrow, darkblue] (env) -- (plan);
\node (tracks) [right of=env, xshift=0.2cm, yshift=0.1cm] {\tiny tracks};

\node[right of=plan, xshift=-.18cm] (arrow-plan-1) {};
\draw [arrow, darkblue] (plan) -- (sc1);
\node (trajec) [right of=arrow-plan-1, xshift=-0.62cm, yshift=0.1cm] {\tiny trajec-};
\node (tory) [right of=arrow-plan-1, xshift=-.7cm, yshift=-0.1cm] {\tiny tory};

\draw [arrow] (mcm) -- (sc2);

\node[right of=sc2,xshift=-0.19cm] (arrow-tr-1) {};
\node[right of=arrow-tr-1, xshift=0cm] (arrow-tr-2) {};
\draw [arrow] (arrow-tr-1) -- (arrow-tr-2);
\node (trajec-1) [right of=arrow-tr-1, xshift=-0.5cm, yshift=0.1cm] {\tiny trajectory};

\node[below of=cav, xshift=-.35cm, yshift=0.75cm] (dotted-1) {};
\node[right of=dotted-1, xshift=7.45cm] (dotted-2) {};
\draw[dashed, thick] (dotted-1) -- (dotted-2);
    
\end{tikzpicture}
    \caption{Data flow between the different components of the function offloading use case service-oriented trajectory planning.}
    \label{dataflow}
\end{figure}
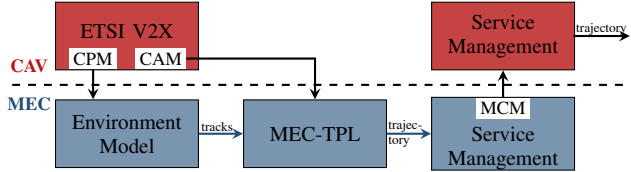

As mentioned before, the service management instances of \ours need to manage the data flow for the corresponding services.
The trajectory planning framework takes as input an environment model and the ego motion of the CAV.
The information about the ego motion can be created using the CAM.
The environment model of the CAV is sent to the MEC server using the ETSI ITS Collective Perception Message~(CPM)~\cite{etsi_tr_103_562_intelligent_2019}.
Another benefit of trajectory planning on the MEC server is that the server can directly benefit from the extended environment model through infrastructure sensors.
The trajectories calculated by the MEC-TPL are received by the SP instance and then encoded into a custom Maneuver Coordination Message~(MCM) that was proposed in~\cite{mertens_21}.
We make use of our custom MCM, because the MCM is currently not yet standardized.
Specifically, the trajectory is encoded into the \textit{SuggestedManeuverContainer} of the MCM from~\cite{mertens_21} and sent to the CAVs. 
The service management instance on the CAV receives the MCMs, extracts the trajectories and then forwards them to the vehicle control system.
The particular data flow of the ETSI ITS messages during activated function offloading for service-oriented trajectory planning is visualized in Fig.~\ref{dataflow}.

\subsection{Quality of Service Requirements}
Trajectory planning is one of the last steps in the execution chain of vehicle automation.
The subsequent control system actuates the vehicle control, such as adjusting the steering angle or acceleration.
Consequently, trajectory planning has strict QoS requirements.
Thus, for the service-oriented trajectory planning, we use a maximum latency $l_\text{max}$ and a maximum inter-arrival time $\Delta t_\text{max}$ as QoS requirements.
The latency is the difference of the trajectory generation time $t_\text{creation}$ to the time when the trajectory is received $t_\text{receiving}$.
The inter-arrival time is defined as the time between two consecutively received trajectories $t_i$ and $t_{i+1}$, $i\in\mathbb{N}_0$.
If the two requirements summarized with
\begin{align}
\begin{aligned}
    l_\text{max} & \leq t_\text{receiving} - t_\text{creation} \\
    \Delta t_\text{max} & \leq t_{i+1} - t_{i}
\end{aligned}
\end{align}
are not fulfilled, the CAV-TPL is re-activated and the SR communicates the termination of the service to the SP.

\section{Evaluation}
We evaluate \ours both in simulation and in a real-world scenario with the considered use case of service-oriented trajectory planning.
\ours is implemented as a ROS2 node in our autonomous driving stack consisting of multiple nodes for perception and planning tasks and a V2X module. The dynamic node handling is achieved through the concept of ROS2 lifecycle nodes.

\subsection{Evaluation in Simulation} 
In the simulation, we compare the application of \ours for service-oriented trajectory planning.
\subsubsection{Simulation Setup}
We use a ROS2-based software-in-the-loop~(SIL) simulation framework that was proposed in~\cite{9636423}.
The SIL simulation runs on Ubuntu 22.04 on an AMD Ryzen 9 7950x CPU with $64\,\unit{GBs}$ RAM.
Each simulated CAV allocates its designated ROS2 nodes of the autonomous driving stack including the \ours node.
The nodes for the simulated MEC server run on separate hardware with an AMD Ryzen 9 3950X CPU with $64\,\unit{GBs}$ RAM.
The two machines are in two distinct local networks that are bridged through a fiber connection. 
The multi-hop connection aims to better represent the network conditions in a wireless connection.
The sending of offloading messages and V2X communication are done using an AMQP broker~\cite{amqp2012} on the machine of the simulated MEC server and different clients for each simulated CAV and the MEC server.

\begin{figure}[t]
    \centering
    \includegraphics[width=\linewidth]{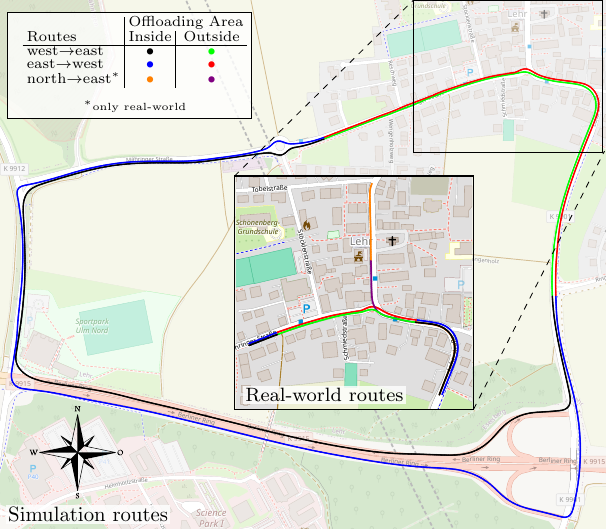}
    \caption{Real-world street layout with the pilot junction and the two round-trips that were emulated in the simulation in the large picture. The zoomed detailed picture shows the same junction with the routes of the real-world tests. Both pictures use the same color coding for the routes given in the legend. Map from~\cite{OpenStreetMap}.}
    \label{fig:street}
\end{figure}

The simulation results involve multiple scenarios with varying number of CAVs ranging from one to six.
The street topology of the simulation is equivalent to a real-world street layout in Ulm-Lehr, Germany, with two round-trips of $\sim 2.9\,\unit{km}$ each.
The equivalent street layout contains a pilot test site that was outlined in~\cite{buchholz_22}.
The street layout for both the simulation and the real-world experiments is shown in Fig.~\ref{fig:street}, where the background shows the round-trips of the simulation and the zoomed traffic junction in the middle the routes for the real-world analysis.
The color change shows the different offloading areas, respectively.
It can be seen that the areas for the simulations are comparatively larger.

We evaluate different scenarios, where the spawn position and route of the involved CAVs on the round-trips are changed to account for different timing conditions of the requests.
For the CODM, we choose the offloading area in a way that $\sim700\,\unit{m}$ of each round-trip is included in the offloading area on the simulated MEC server and an offloading radius of $r_\text{off}=300\,\unit{m}$ for the LODM of the CAVs.
In the simulations, the CAVs drive the round-trip multiple times.
The size of the offloading area results in $\sim\frac{1}{4}$ of the overall length of the round-trip.

\subsubsection{QoS Requirements}
\begin{figure}[t]
    \centering
    \includegraphics[width=\linewidth]{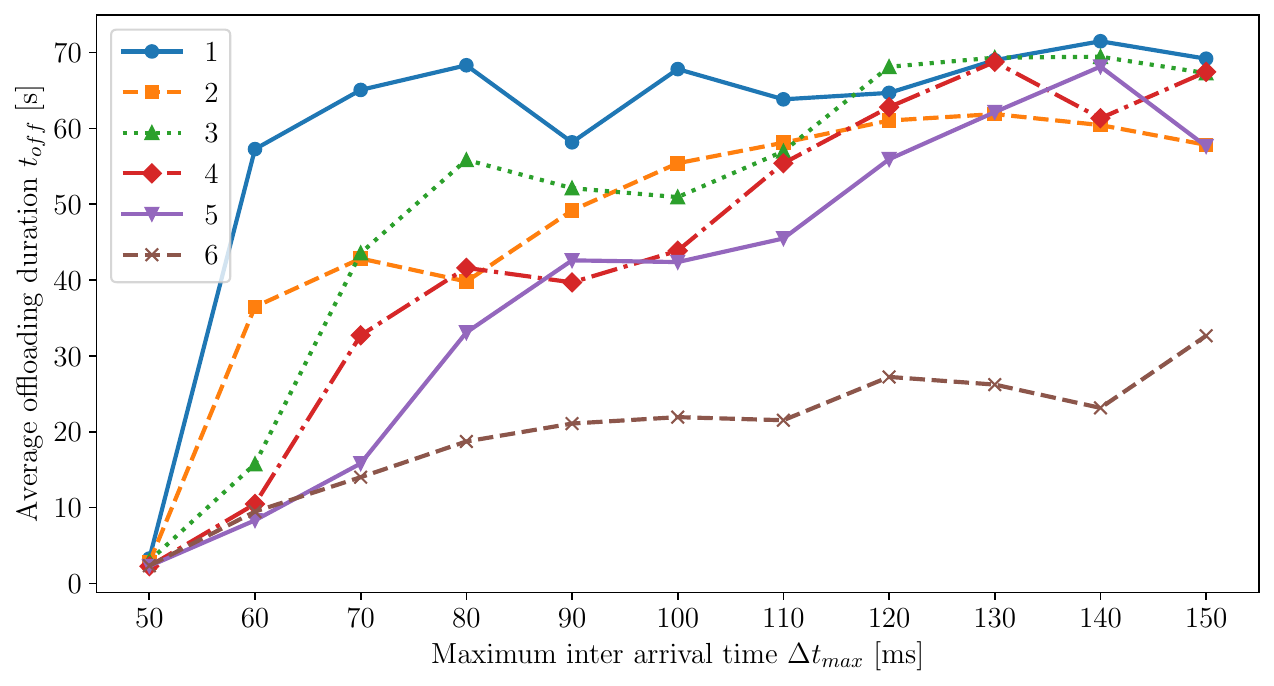}
    \caption{Average offloading duration $t_\text{off}$ depending on the maximum inter-arrival time $\Delta t_\text{max}$ with scenarios of varying number of CAVs.}
    \label{fig:latency-plot}
\end{figure}
To determine suitable parameters for the QoS requirements, we first analyze the influence of different parameters on different scenarios.

Fig.~\ref{fig:latency-plot} shows the offloading duration $t_\text{off}$ of different scenarios involving a different number of vehicles depending on the maximum allowed inter-arrival time $\Delta t_\text{max}$.
$t_\text{off}$ is the overall time of how long the trajectory planning service is offloaded for each accepted service request.
We ran each scenario with varying CAV count $\in[1,\ldots,6]$ and $\Delta t_\text{max}\in[10,\ldots,150]\unit{ms}$ for 30 minutes, thus conducting a total of 33 hours of simulation, including numerous round-trips for each simulated CAV.
With higher $\Delta t_\text{max}$, more trajectories can be received within the time limit, which means that offloading is possible for a longer time.
This can also be seen in Fig.~\ref{fig:latency-plot}, where the offloading duration $t_\text{off}$ generally increases with higher $\Delta t_\text{max}$. 
The maximum offloading duration of the scenario is bounded to $\sim70\,\unit{s}$, which corresponds to the time the CAVs spend in the offloading area.

Fig.~\ref{fig:latency-plot} also shows a slower increase of $t_\text{off}$, and thus a generally shorter offloading duration for more vehicles.
Especially for scenarios with six CAVs, the maximum offloading duration is lower than for less complex scenarios.
The first reason for this is the higher network congestion and workload on the MEC server with more offloading requests resulting in more violations of the QoS requirements, and thus an earlier termination of the offloading.
The second, however, results from the simulation setup.
With six vehicles present, the machine running the simulation is close to its computational limits, which also reduces the offloading duration for these scenarios.
Thus, the network congestion is not solely responsible for the worse performance and a more profound simulation setup might improve the results.

\begin{figure}[t]
    \centering
    \includegraphics[width=\linewidth]{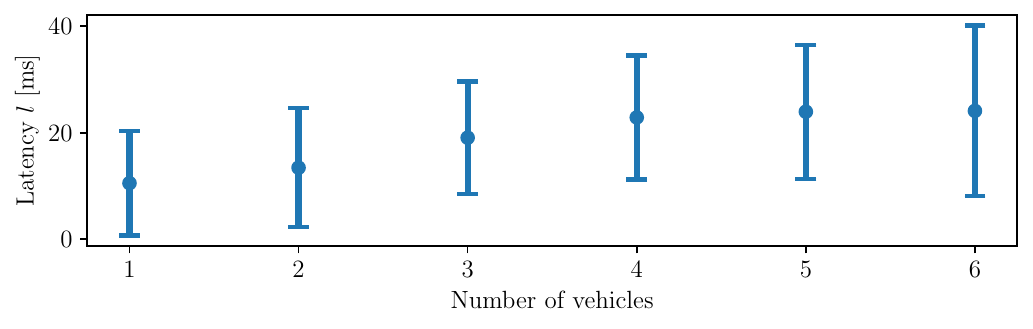}
    \caption{Mean latency values including the standard deviation depending on the number of simulated vehicles.}
    \label{fig:delay-plot}
\end{figure}
For the analysis shown in Fig.~\ref{fig:latency-plot}, the second parameter of the QoS requirements, i.e., the maximum latency, is set to $l_\text{max}=50\,\unit{ms}$.
To assess the validity of this value, we stored the latencies of all received trajectories of the simulations.
Fig.~\ref{fig:delay-plot} shows the mean delay values and its standard deviations depending on the number of simulated vehicles in the scenarios.
It can be seen that the both the mean latency and the standard deviation increase with higher number of vehicles.
This is a result of the higher and more varying network congestion following from more vehicles, respectively.
The distribution also shows that, with the given setup, it is highly probable that the latency of a received trajectory is less than $50\,\unit{ms}$.

Considering the real-time requirements for vehicle control and from the results in Fig.~\ref{fig:latency-plot}, we determine as suitable QoS requirements:
\begin{align}\label{qos-requirements}
\begin{aligned} 
    \Delta t_\text{max}=100\,\unit{ms} \\
    l_\text{max}=50\,\unit{ms}
\end{aligned}
\end{align}

\subsubsection{CPU Usage}
One of the main motivations of function offloading was to reduce the energy consumption of the battery-powered CAVs.
For that, we evaluate the average CPU usage in percent in relation to the capacity of one core of the CPU of the corresponding systems with function offloading activated and deactivated, since needed computational power correlates with the energy consumption.
It is worth mentioning that the overall CPU usage of all CAVs and the MEC server is evidently higher due to the additional functionality.
However, the main focus is the minimization of the hardware consumption of the CAVs where we assume that we have limited power available.

On the CAV, the overall CPU usage without and with \ours are characterized by
\begin{align}\label{eq:cpu-usage}
\begin{aligned}
    c_\text{without} &= t_{\text{total}}\cdot c_{\text{TPLa}} \text{\ \ and} \\
    c_\text{with} &= t_{\text{total}}\cdot c_{\text{\ours SR}} + t_a c_{\text{TPLa}} + t_d\cdot c_{\text{TPLd}},
\end{aligned}
\end{align}
respectively, with $t_{\text{total}}$ the total time and $t_d$, $t_a$ the times with TPL deactivated and activated, respectively. 
$c_{\text{TPLa}}$, $c_{\text{TPLd}}$, and $c_{\text{\ours SR}}$ are the respective average CPU percentages for TPLa, TPLd, and \ours as a SR. In another simulation, where we set the QoS requirements as in Eq.~\eqref{qos-requirements}, we recorded the average CPU usage values, which can be seen in the second column of Table~\ref{table:cpu-usage}.

With the main goal being the reduction of CPU usage through function offloading, the objective can be formulated as
\begin{align}\label{eq:obj}
    c_\text{with} < c_\text{without}.
\end{align}

\begin{table}
\centering
\caption{Average CPU usage of relevant systems w.r.t one CPU core.}
\begin{tabular}{c | c | c}
    System & Simulation & Real-World \\
    \hline
    TPLa $c_\text{TPLa}$ & $19.5\%$ & $116.1\%$ \\
    TPLd $c_\text{TPLd}$& $8.5\%$ & $28.3\%$ \\
    \ours SR $c_\text{\ours SR}$ & $0.96\%$ & $2.61\%$
\end{tabular}
\label{table:cpu-usage}
\end{table}
When inserting Eq.~(\ref{eq:cpu-usage}) into Eq.~(\ref{eq:obj}) and with $t_a = t_\text{tot} - t_d$, the proportion of how long offloading should be activated w.r.t the total application time to reach the objective is
\begin{align}\label{ratio}
    \frac{t_d}{t_\text{tot}} > \frac{c_\text{\ours}}{c_\text{TPLa} - c_\text{TPLd}} = 0.0793 \text{\ \ }\widehat{=}\text{\ \ } 7.93\,\%,
\end{align}
with $0.0793$ resulting from the simulated values in Table~\ref{table:cpu-usage}.

In our experiments, we also analyzed the CPU usage of the V2X system that is used to send the CAMs, CPMs, and MCMs from the CAV to the MEC server and vice versa.
However, when assuming that CAMs and CPMs are sent even without function offloading for other V2X functionality, the additional usage due to function offloading itself is negligible and thus excluded from the evaluation.

It can be concluded that in the simulation setup, SOFOF can improve the overall CPU workload of the CAVs for service-oriented trajectory planning, if the trajectory planning task is offloaded for at least $7.93\%$, with great improvements for higher offloading ratios.

\subsection{Real-World Evaluation}\label{rwe}
Since the latency considerations in the simulation do not necessarily reflect the application of the framework in a real-world scenario, we also deploy \ours on our autonomous test vehicle and on the MEC server of our test site~\cite{buchholz_22}.
We evaluate the adherence of the QoS requirements in a similar way as in the simulation, showing the robustness towards network effects resulting from the dynamic nature of real-word traffic scenarios.

\subsubsection{Real-World Setup}
We again use ROS2 on the test vehicle and the MEC server, respectively.
The autonomous test vehicle is a Mercedes-Benz E-class equipped with an AMD Ryzen Threadripper 3990X CPU with $128\,\unit{GBs}$ RAM, while the MEC server comprises an AMD Ryzen Threadripper PRO 5965WX CPU with $178\,\unit{GBs}$ RAM.
The network connection is established through a 5G network with an AMQP connection to handle the V2X communication.
Derived from the simulation, we set the QoS requirements to $\Delta t_\text{max}=100\,\unit{ms}$ and $l_\text{max}=50\,\unit{ms}$.

For safety reasons, we defined a different offloading area for the CODM on the MEC server compared to the simulations.
In particular, the chosen offloading area includes $\sim220\,\unit{m}$ of the predefined routes in the vicinity of our pilot traffic junction, see Fig.~\ref{fig:street}.
The actual possible area could be much larger.
We conducted $15$ test drives on three different routes through the area shown in Fig.\ref{fig:street}, where the \ours algorithm dynamically changes the trajectory planning service from CAV-TPL to MEC-TPL depending on the position of our autonomous test vehicle and the defined offloading area.
Similarly to the simulation analysis, we monitored the needed CPU usage of the relevant ROS2 nodes, i.e., trajectory planning (TPLa and TPLd) and function offloading (\ours SR), while we excluded SOFOF SP since it runs on the MEC server with a wired energy supply.

\subsubsection{CPU Usage}
The third column of Table~\ref{table:cpu-usage} shows the average observed CPU usage of the different components of the autonomous test vehicle.
When comparing the two columns for the simulation and real-world analysis, it is noticeable that the CPU usage of TPLa is significantly higher in the real-world evaluation than in the simulation for the one vehicle.
This is a result of a different parameter setup of the TPLa, where the update frequency of the MPC planning algorithm, see~\cite{10186535}, is set to $100\,\unit{Hz}$ for the real-world application instead of $20\,\unit{Hz}$ in the simulation.
The higher update frequency ensures safer trajectory planning, as it can adapt faster to changes in the environment of the vehicle.
It is worth noting that the QoS requirements are not adjusted, since the chosen values are still valid as a fallback solution if the latency and inter-arrival time are not met.
Also, since we take the percentage values in relation to one CPU core, the value can be $>100\%$, meaning that the process needs multiple CPU cores to run.

The implementation of \ours on the real-world test vehicle also needs a higher percentage of the CPU than in the simulation.
The reason for this is similar to the higher value of TPLa.
Since the published trajectories need to be handled by \ours, the CPU usage is consequently higher for a higher publish frequency.

With the objective from Eq.~\eqref{eq:obj}, the time ratio for how long the trajectory planning needs to be deactivated in order for the system with \ours to use less of the CPU is
\begin{align}
    \frac{t_d}{t_\text{tot}} > 0.0615\text{\ \ }\widehat{=}\text{\ \ } 6.15\,\%.
\end{align}

We also recorded the latency for each trajectory that was sent from the MEC server to our test vehicle. For the real-world scenarios, the mean and standard deviation values were
\begin{align}\label{eq:delay-real}
\begin{aligned} 
    l_\text{avg}^\text{real}&=11.94\,\unit{ms} \\
    l_\text{std}^\text{real}&=2.98\,\unit{ms}.
\end{aligned}
\end{align}

Especially the value for the standard deviation from Eq.~\eqref{eq:delay-real} varies with the values for one simulated scenario shown in Fig.~\ref{fig:delay-plot}.
In particular, for one vehicle, Fig.~\ref{fig:delay-plot} shows $l_\text{avg}^\text{sim,1}=10.54\,\unit{ms}$ and $l_\text{std}^\text{sim,1}=9.83\,\unit{ms}$.
One reason for the lower standard deviation is that in the real-world analysis, only the data relevant for function offloading is sent to the server, while in the simulation additional simulation data needs to be sent to the machine simulating the server.

With the observed values $l_\text{avg}^\text{real}$, $l_\text{std}^\text{real}$ and $l_\text{max}=50\,\unit{ms}$, a very high percentage of trajectories is received within the maximum delay.
Nevertheless, in $2$ of the in total $15$ conducted scenarios, the function offloading had to be terminated early due to a violation of $l_\text{max}$ within the offloading area shown in Fig.~\ref{fig:street}.
However, the dynamic switch back to the CAV-TPL resulting from the violation was not noticeable in the driving experience of the passengers, demonstrating the smooth and safe transitions between the trajectory planning services even under varying network conditions.

Therefore, it can be stated that real-world experiments not only support the simulation results but also show that the application of \ours can efficiently extend the local computational hardware and software of our autonomous test vehicle.

\section{Conclusion and Future Work}
This paper presented \ours, a generic function offloading framework in the context of SOAs that can extend the number of available services for the task of autonomous driving.
Exemplarily, the task of offloading the trajectory planning was introduced as a use case and was handled by the framework.
In the evaluation, the benefit of \ours in improving hardware usage on the CAVs was shown in simulation.
The real-world evaluation showed the applicability of the framework to actual scenarios, showing that function offloading holds great promise for the development of future mobility systems.

Due to the generic design, \ours opens a wide range of further research.
Our future work will focus on enhancing the system to further benefit from function offloading by involving multiple different services including cooperative functions, e.g., cooperative intersection management~\cite{9827300}.
We will also focus on enhancing the simulation setup to better evaluate the scalability of \ours and consider the aspect of safety by not only using the QoS requirements, but also analyzing the quality of the offloaded tasks through analyzing the downstream data of the different services.

\addtolength{\textheight}{-3cm}   



\bibliographystyle{IEEEtran}
{ \footnotesize \bibliography{FuncOff} }

\end{document}